\pdfoutput=1
\documentclass[conference]{IEEEtran}
\usepackage{arabtex}
\usepackage{cite}
\usepackage{utf8}
\usepackage{times}
\usepackage{url}
\usepackage{latexsym}
\usepackage{graphicx}
\usepackage{fancyhdr}
\usepackage{hyperref}
\fancypagestyle{firstpage}{% Page style for first page
  \fancyhf{}% Clear header/footer
  
  \fancyfoot[L]{{\em This paper was published in the Proceedings of the 2017 IEEE 11th International 
Conference on Semantic Computing (ICSC), pages 17-23, San Diego, CA, USA, January 2017. \copyright 2017 IEEE\\
Link to article abstract in IEEE Xplore: http://dx.doi.org/10.1109/ICSC.2017.75}}% Footer
}
\setcode{utf8}

\title{Filtering Tweets for Social Unrest}

\author{\IEEEauthorblockN{Alan Mishler}
\IEEEauthorblockA{Department of Statistics\\
Carnegie Mellon University\\
Pittsburgh, PA 15213\\
amishler@andrew.cmu.edu}
\and
\IEEEauthorblockN{Kevin Wonus}
\IEEEauthorblockA{Data Science Unit\\
The DarkStar Group\\
St. Petersburg, FL 33701\\
kevin@wonus.com}
\and
\IEEEauthorblockN{Wendy Chambers}
\IEEEauthorblockA{CASL\\
University of Maryland\\
College Park, MD 20742\\
wchamber@umd.edu}
\and
\IEEEauthorblockN{Michael Bloodgood}
\IEEEauthorblockA{Department of Computer Science\\
The College of New Jersey\\
Ewing, NJ 08628\\
mbloodgood@tcnj.edu}
}

\date{}

\begin{document}

\pagenumbering{gobble}

\maketitle

\thispagestyle{firstpage}% firstpage page style for first page

\begin{abstract}
Since the events of the Arab Spring, there has been increased interest in using social media to anticipate social unrest. While efforts have been made toward automated unrest prediction, we focus on filtering the vast volume of tweets to identify tweets relevant to unrest, which can be provided to downstream users for further analysis. We train a supervised classifier that is able to label Arabic language tweets as relevant to unrest with high reliability. We examine the relationship between training data size and performance and investigate ways to optimize the model building process while minimizing cost. We also explore how confidence thresholds can be set to achieve desired levels of performance.
\end{abstract}
\section{Introduction} \label{introduction}

There has been substantial interest in building technologies that can use social media
postings to help forecast civil unrest
\cite{compton2014,muthiah2015,ramakrishnan2014}. The Arab Spring of 2011 compellingly illustrates how social
media can both reflect and influence political (in)stability \cite{campbell2011}. 

Since social media data is generated on such a large and rapid scale,
computational tools are potentially extremely useful in helping to render meaning from that data.  
While previous work has focused on forecasting specific near-term unrest events
\cite{muthiah2015}, in this current paper we are interested in filtering social media
content for postings that are relevant to social unrest, with the idea that downstream
systems or human experts would use this filtered content for further analysis.  

In particular, we experiment with filtering tweets written in Arabic for relevance
to social unrest. To do so, we frame the problem as a text classification problem with
two classes: relevant to social unrest and not relevant. We experiment with creating
annotated data and using machine learning to build a social unrest relevance classifier.
We use multiple data representations and multiple inference algorithms. We find that a
bag of words representation with an SVM learning algorithm works quite well. 

Since annotation is expensive and time-consuming, we investigate data sizes needed to
achieve various levels of performance, and we investigate the utility of active learning for learning
stronger classifiers with less data. To realize the potential gains that active learning
enables, we explore the use of automated annotation stopping methods and find they are
effective for this application as well. 

Downstream consumers of our system's filtered tweets, whether automated systems or
people, will have different preferences for overall performance levels, and
precision-recall tradeoffs in particular. Accordingly, we also investigate how changing
the system confidence threshold for our social unrest relevance filter impacts these
performance metrics. 

Section~\ref{relatedWork} discusses related work. Section~\ref{dataAndTaskDescription} describes the data we used for our experiments
and describes our filtering task in more detail. Section~\ref{experimentalSetup}
provides details regarding our experimental setup. Section~\ref{dataSizeExperiments}
presents the results and analyses from our data size experiments, including our active
learning results and annotation stopping detection results.
Section~\ref{confidenceExperiments} presents the results and analyses from our system
confidence experiments, and section~\ref{conclusion} concludes.

\section{Related Work} \label{relatedWork}

An abundance of systems have been built in recent years to produce structured predictions of unrest events. One approach applies message enrichment and a series of machine learning models to tweets to forecast the location, date, involved population, and event type of unrest events \cite{ramakrishnan2014}. Others use a cascade of text filters to directly extract tweets \cite{compton2014} or Tumblr posts \cite{xu2014} that identify planned unrest events. Another group retrospectively examines the Egyptian revolution of 2011, using Twitter user metadata, including graph features of user networks, to predict major events \cite{boecking2015}.

In addition to systems designed to predict major events, a number of researchers have examined the problem of identifying tweets related to predetermined ongoing events. For example, some have compared tweets to news stories about specific events, using clustering to identify tweets relevant to those events \cite{hua2013}. Others use linguistic features to identify the geographic origin of non-geotagged tweets, in order to identify tweets coming from within regions experiencing a crisis or emergency \cite{morstatter2014}.

Finally, quite a few systems use SVMs to classify tweets along dimensions such as sentiment \cite{jumadi2016}, political alignment \cite{conover2011}, categories of news \cite{dilrukshi2013}, relevance to influenza \cite{aramaki2011}, etc.

Broadly speaking, there is a distinction between systems which aggregate information from tweets or use tweets to produce predictions, and systems which filter tweets to identify those relevant to a particular purpose or goal. (Some systems, like \cite{compton2014}, do the latter by means of the former.) Here, we consider the problem of identifying tweets in Arabic that are relevant to social unrest, for use in downstream analysis. The downstream user may be a computational system like the ones described above or may be a human analyst. We assume that decisions designed to prevent or respond to social unrest are made by humans, who may wish not only to predict specific unrest events but to gain insight into the causes and conditions surrounding unrest.

There are myriad research questions related to social unrest which computers are not able to answer in an automated fashion. For example, researchers may wish to study the linguistic features of relevant tweets, gain insight into the actors involved, or form narratives to explain unrest events. Linguists, psychologists, and sociologists can incorporate cultural and historical knowledge that computers are as yet unable to represent. Tweets may discuss conditions likely to lead to unrest, such as tensions between ethnic groups or complaints about the government, without alluding to planned unrest events. 

Unlike the systems described above, we do not aim to predict short-term unrest events or identify tweets relevant to specific, predetermined events. Instead, we aim to identify tweets that are broadly relevant to social unrest and political instability, ranked by the likelihood of relevance. The filtered tweets could be used to predict unrest events, or they could be used by human analysts to address a variety of other research questions.
\section{Data/Task Description} \label{dataAndTaskDescription}

For the purposes of annotation, we define social unrest as the public expression of discontent, including public 
protest that does not threaten the regime's hold on power, and/or sporadic but low-level violence. 

We were provided with tweets collected using the Twitter API and a keyword list consisting of 709 Arabic terms related to social unrest 
that was developed by an Arabic lexicographer and other members of the research team fluent in Arabic. The keywords were treated as being joined by a Boolean OR, so tweets were returned that contained one or more of the keywords. The keyword list was designed to be durable and usable across different countries and conflicts, so it contained only high frequency, non-proper nouns in Modern Standard Arabic, such as ``protest,'' ``police,'' and ``assassinate.'' The terms were drawn from police and military lexicons or were derived from Google searches by members of the research team. Terms were avoided which had been previously found to occur primarily in contexts not related to unrest. For example, the word \<جمهور> \textit{jumhuur} ``masses (of people)'' was not included because it was primarily found in tweets related to soccer matches and movie stars.

The Arabic terms were morphologically expanded in order to provide coverage 
for their most frequent word forms. 
For example, the word \<متنازع> \textit{mutanazie} ("``conflicting/clashing,''€) was represented in the morphologically 
expanded forms \<المتنازع> \textit{al-mutanazie} ("``the-conflicting/clashing,''" masc.) and \<المتنازعة> \textit{al-mutanaziea} ("``the-conflicting/clashing,''" fem.). 
A total of 33,922,037 tweets were collected between July, 2015 and April, 2016. Duplicate tweets, including retweets, were removed, as were tweets containing 
nonstandard Arabic characters (e.g. in Urdu and Farsi) or non-Arabic characters (e.g. Bopomofo, Hangul, Hiragana, Katakana). Tweets were also removed that contained pornographic keywords designed to direct readers to porn sites. After cleaning, 16,165,081 
tweets remained, from which tweets were sampled for annotation in batches of approximately 1,000, stratified by timestamp.

Annotation took place in two rounds over the the course of 21 weeks, not including training periods. In round one, two annotators were selected who had been the top performers in terms of reliability and quality in a previous 
annotation effort. Both were fluent in English and Arabic. After week 12, a third annotator who was fluent in English and Arabic was added in order to increase annotation capacity, at which point all three annotators received three weeks of additional training (due to a pause of several months in annotation between the rounds). 

The annotators received a detailed introduction to the annotation guide, which defined social unrest, provided background on unrest in the Arabic-speaking world, and gave examples of tweets considered relevant or irrelevant to social unrest. Each week during the training periods, the annotators coded the same sets of approximately 100 tweets. Weekly 2-3 hour ``consensus'' meetings were held to discuss the annotation process and to identify and resolve any disagreements. 

During the formal annotation periods, approximately 10-13\% of the tweets were given to all the annotators for the purposes of 
establishing inter-rater reliability. Weekly consensus meetings were held to discuss and resolve discrepancies in the annotation of the shared tweets. Reliability between the first two annotators was measured by Cohen's kappa \cite{cohen1960} and percent agreement \cite{fleiss1971}. Reliability among the three annotators in the second round of annotation was determined using two-way mixed intraclass correlation (ICC) with absolute agreement (model 3) \cite{shrout1979}. Reliability was high throughout and improved over the course of annotation (Table \ref{t:interrater}).

\begin{table*}[ht]
\small
\begin{center}
\begin{tabular}{|l|c|c|c|}
\hline 
& \multicolumn{2}{c}{\bf Round 1} & \bf Round 2 \\
\hline
& \bf Cohen's $\kappa$ & \bf \% agreement & \bf Intra-class correlation \\
First week & 0.61 & 0.87 & 0.87 \\
Final week & 0.82 & 0.92 & 0.90 \\
Mean & 0.67 & 0.86 & 0.89 \\
\hline
\end{tabular}
\end{center}
\small
\caption{\label{t:interrater} Inter-rater reliability scores.}
\end{table*}

Over the course of annotation, an additional 1\% of the tweets were removed 
because they were non-Arabic or of a pornographic nature. In total, 21,711 tweets were annotated for relevance 
to social unrest (1 = relevant, 0 = irrelevant). 13,535 of those tweets (62\%) were deemed relevant.

\section{Experimental Setup} \label{experimentalSetup}

In this section we describe our experimental setup. Subsection~\ref{preprocessing} explains our preprocessing steps, 
and subsection~\ref{dataRepresentation} describes our data representations. 

\subsection{Preprocessing} \label{preprocessing}

In order to preserve emojis in the tweets during subsequent processing, a table of emojis was used to replace all 
matching emojis in the tweets with the string emojiX, where X was an index between 1 and 842. The table of emojis was 
assembled from \cite{whitlock2015}. The tweets were then tokenized using MADAMIRA, an open source Arabic parser that is 
available online \cite{pasha2014}. MADAMIRA contains parsers for Modern Standard Arabic (MSA) and Egyptian Arabic; we parsed the tweets using the MSA parser.

Since Arabic is morphologically complex, we lemmatized the tokens in order to avoid a proliferation of features based on morphological variants of similar words. We used MADAMIRA both to lemmatize each token and to label the Part of Speech (POS) for each token, using the 
ATB4MT tokenization and parsing scheme. (Hashtags were left intact, not lemmatized. See the MADAMIRA user manual for details.) MADAMIRA uses the Buckwalter 
POS tagset for Arabic, of which there are 54 basic components that can be combined with inflectional 
markers for categories such as person, gender, number, and case to produce a much larger 
tag set\footnote{Buckwalter, T. (2002). Buckwalter Arabic Morphological Analyzer Version 1.0. Linguistic Data Consortium, 
University of Pennsylvania. LDC Catalog No.: LDC2002L49}. 
MADAMIRA also includes an additional component, FUNC\_WORD. We collapsed these 55 basic components into 19 simple POS tags. For example, tags for different kinds of adjectives---comparative, ordinal, deverbal, and proper adjectives---were collapsed into a single ADJ tag.
Both the lemmas and the POS tags were then recombined separately 
into the original order, so that each tweet consisted of (1) lemmatized versions and (2) POS versions of the original tokens in the tweet.

Finally, we cleaned the lemmatized output. We removed newlines and all punctuation except for \#, @, and underscores (in order 
to retain hashtags and Twitter usernames, including user mentions). We converted all URLs to a token ``LINK'' and converted all numbers that 
were not attached to text strings to a token ``NUMBER.'' We also removed stop words. Our stop word list was assembled by 
aggregating the 1,000 most frequent words in the arabiCorpus, a free Arabic corpus of approximately 
173 million words\footnote{arabicorpus.byu.edu}. A researcher fluent in Arabic then removed all content 
words (mostly nouns, adjectives, and names), leaving behind a set of 651 stop words.

\subsection{Data Representation} \label{dataRepresentation}

We created six bag-of-ngrams feature representations using the lemmatized and POS versions of the tweets. 
Our features were binary (1 = present within the tweet, 0 = not present.) The feature sets consisted of POS unigrams (POS1); 
POS unigrams and bigrams (POS1-2); and POS unigrams, bigrams, and trigrams (POS1-3); lemma unigrams (Lex1); 
lemma unigrams and bigrams (Lex1-2); and lemma unigrams and bigrams combined with POS unigrams, bigrams, and trigrams (Lex1-2\_POS1-3). 
Only ngrams that occurred 3 or more times in the data were included. For each feature set, we created a Naive 
Bayes classifier and a Random Forest classifier using Weka \cite{hall2009}, and a Support Vector Machine classifier 
using SVM$^{light}$ \cite{joachims1999}. These models used the first 10,066 tweets, roughly half of the data that 
was ultimately annotated. 6,469 of these tweets (64\%) were relevant. The SVM with Lex1 yielded the best performance 
overall (Table \ref{t:feature_sets}), so we used this model and feature set for all subsequent models. All models were trained using the default learning settings in SVM$^{light}$, including the default C parameter and a linear kernel.

\begin{table*}[ht]
\small
\begin{center}
\begin{tabular}{|l|l|lll|lll|lll|}
\hline \bf Name &\bf  Size & \multicolumn{3}{|c|}{\bf Naive Bayes} & \multicolumn{3}{|c|}{\bf Random Forest} & \multicolumn{3}{|c|}{\bf SVM} \\ \hline
         &   & \bf Prec. & \bf Rec. & \bf F1 & \bf Prec. & \bf Rec. & \bf F1 & \bf Prec. & \bf Rec. & \bf F1 \\
       POS1 & 19 & 0.69 & 0.91 & 0.78 & 0.68 & 0.88 & 0.77 & 0.68 & 0.96 & 0.79 \\
       POS1-2 & 380 & 0.74 & 0.74 & 0.74 & 0.69 & 0.90 & 0.79 & 0.67 & 0.96 & 0.79 \\
       POS1-3 & 7,239 & 0.75 & 0.72 & 0.74 & 0.69 & 0.91 & 0.79 & 0.68 & 0.94 & 0.79 \\
       Lex1 & 6,152 & 0.89 & 0.83 & 0.86 & 0.85 & 0.90 & 0.87 & 0.87 & 0.88 & 0.88 \\
       Lex1-2 & 17,235 & 0.91 & 0.79 & 0.84 & 0.86 & 0.88 & 0.87 & 0.87	& 0.88 & 0.88 \\
       Lex1-2\_POS1-3 & 24,474 & 0.88 & 0.78 & 0.83 & 0.76 & 0.92 & 0.83 & 0.85 & 0.88 & 0.86 \\
\hline
\end{tabular}
\end{center}
\small
\caption{\label{t:feature_sets} Number of features (Size), precision, recall, and F1 for each of six feature sets for each of the three models, estimated using 10-fold cross-validation.}
\end{table*}

\section{Data Size Experiments} \label{dataSizeExperiments}
\subsection{Experimental setup}
Given that human annotation is costly and time consuming, when building a classifier it is important to 
determine when sufficient data has been annotated and to select data for annotation that produces maximum 
performance benefits for the classifier. We investigated the relationship between training data size and performance, 
using active learning. Active learning aims to select unlabeled instances for 
annotation that are likely to provide the greatest marginal improvements in classifier performance. Additionally, 
we investigated an algorithm designed to determine an annotation stopping point, beyond which additional annotation 
provides only negligible improvements in performance.

The data was randomly divided into 10 test folds. For each fold, the remaining nine folds 
were recombined into a training pool. Using an SVM classifier, we performed 10-fold cross-validation 
using two different groups of 100 training sets sampled from this pool. The training sets ranged in size from 356 to 19,540 (the entire training pool). 
The \textbf{Random} training sets consisted of randomly sampled tweets from the training pool. 
The \textbf{Active Learning} training sets were constructed using active learning: a model trained on the 
initial set was used to classify the remaining tweets in the training pool, yielding a set of scores (positive and 
negative real numbers) for each tweet, with the absolute values of the scores representing the Euclidean distance of each tweet 
in feature space from the learned hyperplane. The closest $N$ absolute values (where $N =$  the 
step size between training sets) were added to the training set, a new model was trained, and the process was repeated. 
Since, in some sense, the tweets closest to the class boundary represent the instances that the 
model is least sure about, adding them to the training set is likely to yield greater marginal improvements 
in model performance than random tweets, which may be more similar to tweets the model has already been trained on. 
This method has been successful in past work \cite{bloodgood2009a,campbell2000,schohn2000,tong2002}. 

To determine an annotation stopping point, we used an algorithm based on stabilizing model 
predictions \cite{bloodgood2009b,bloodgood2013}. 
For each test fold $i = 1, \ldots, 10$ and each set of predictions $p_j, j = 2, \ldots, 100$, Cohen's kappa ($\kappa_{ij}$) was 
calculated between $p_j$ and $p_{j-1}$. A stopping point $\kappa_{is}$ was designated if and when three successive kappa values 
were observed at or above a threshold of 0.99, i.e. when $\kappa_{ij} \geq 0.99, j = s-2, s-1, s$. 

\subsection{Results}
\begin{figure}
\includegraphics[keepaspectratio=true, width=\columnwidth]{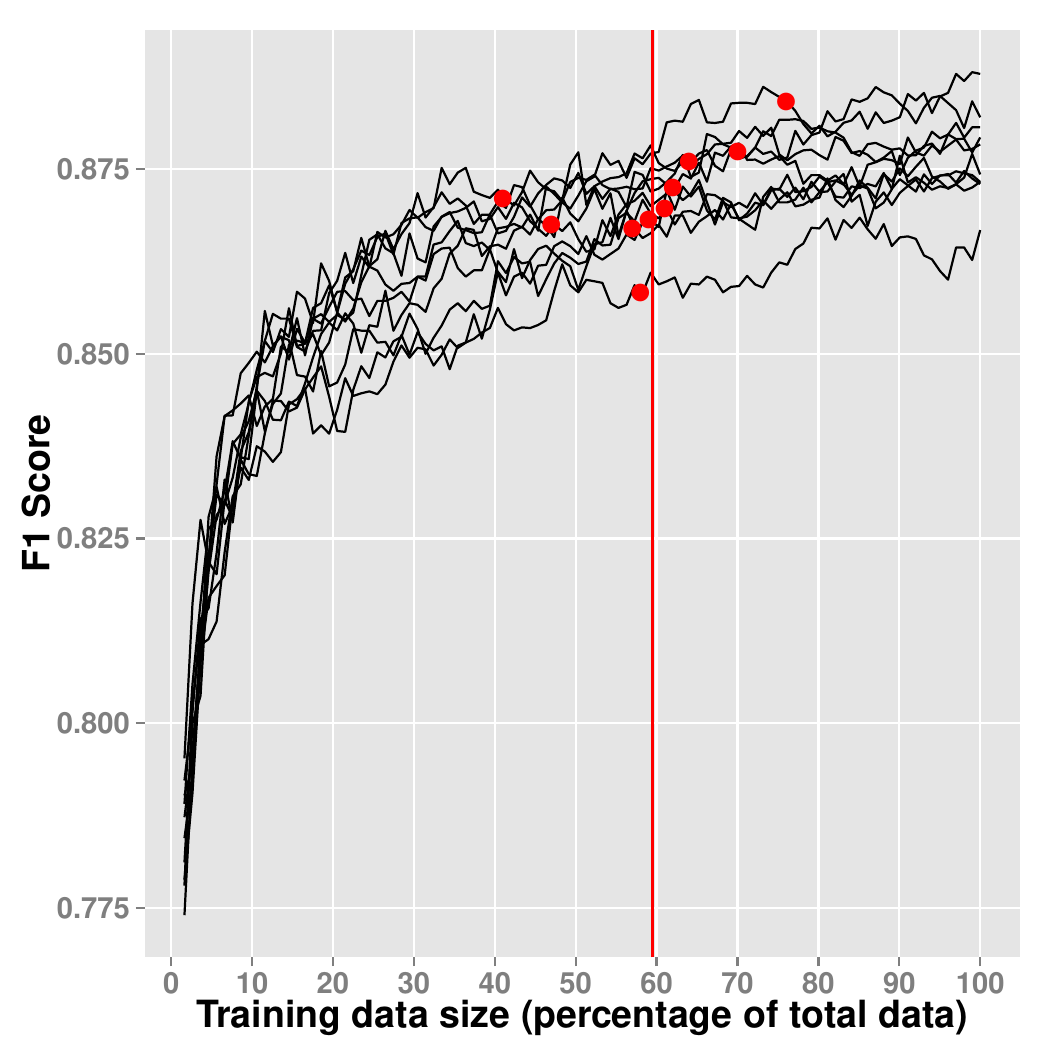}
\caption{Training curves, stopping points, and mean stopping point for the Random models}
\label{f:random_noPA}
\vspace{-.5cm}
\end{figure}

\begin{figure}
\includegraphics[keepaspectratio=true, width=\columnwidth]{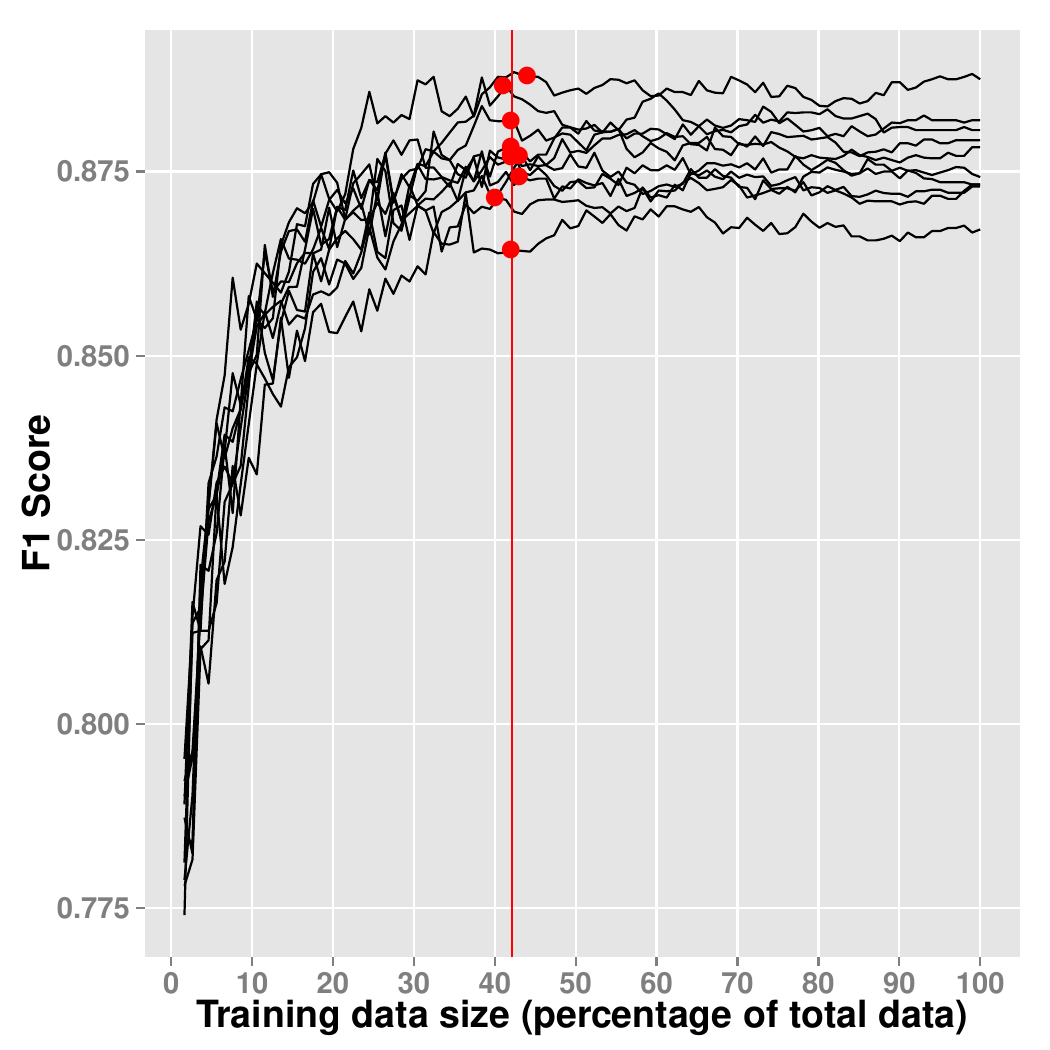}
\caption{Training curves, stopping points, and mean stopping point for the Active Learning models}
\label{f:closest_noPA}
\vspace{-.5cm}
\end{figure}

\begin{figure}
\includegraphics[keepaspectratio=true, width=\columnwidth]{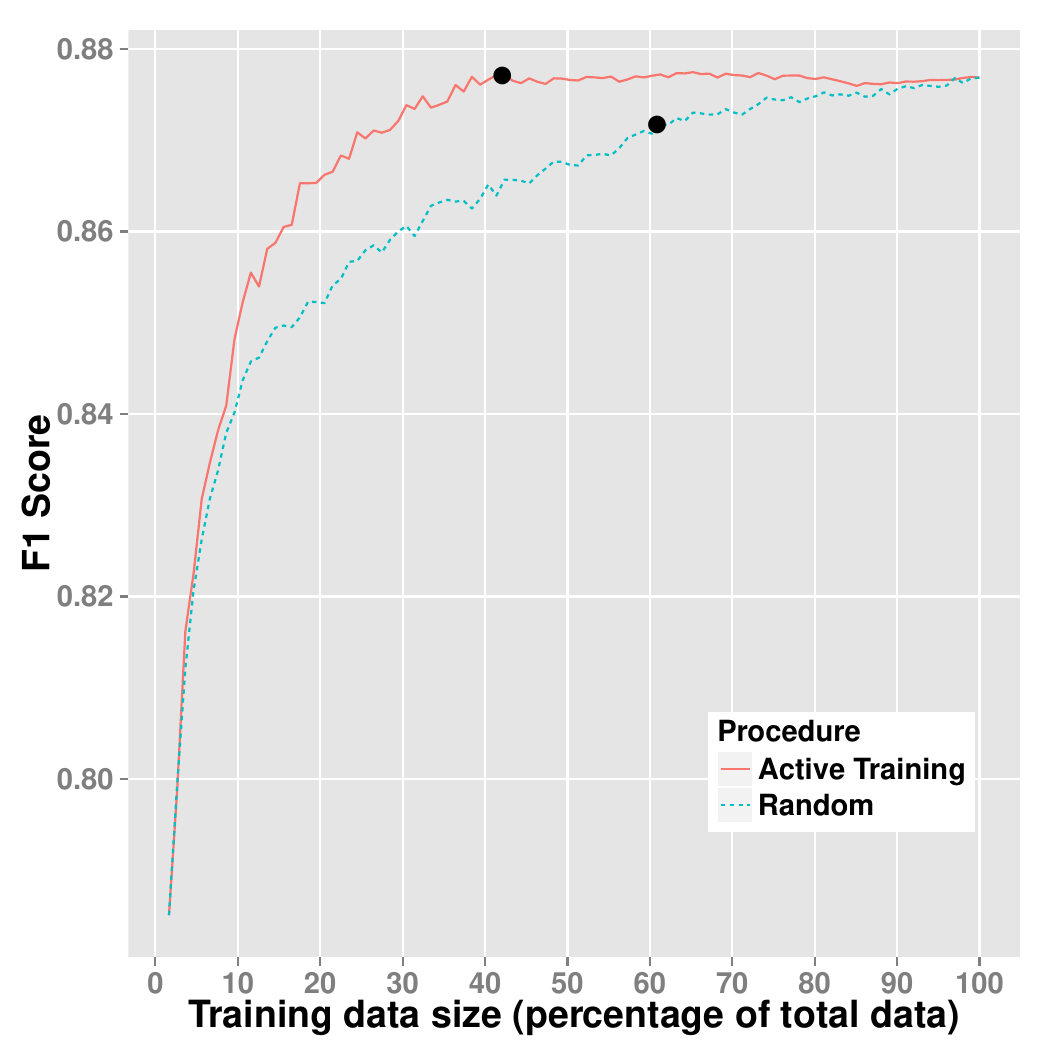}
\caption{Mean training curves and stopping points for the Random and Active Learning models}
\label{f:means_all_algorithms}
\end{figure}

Figures \ref{f:random_noPA} and \ref{f:closest_noPA} show the Random and Active training curves and stopping points for individual folds. Figure \ref{f:means_all_algorithms} shows the mean Random and Active Learning training curves and stopping points. Performance metrics for the final model are given in Table \ref{t:final_metrics}. (Note that the final Random and Active Learning training sets are the same, since they both consist of the entire training pool.)

\begin{table}[ht]
\small
\begin{center}
\begin{tabular}{|l|l|l|}
\hline \bf Precision & \bf Recall & \bf F1 \\ \hline
       0.88 & 0.87 & 0.88 \\
\hline
\end{tabular}
\end{center}
\small
\caption{\label{t:final_metrics} Performance metrics for final model with 10-fold cross-validation.}
\end{table}

As seen in Figure \ref{f:means_all_algorithms}, active learning provides a substantial benefit over random sampling, 
yielding higher F1 scores for a given training data size, and, conversely, requiring smaller training sets to achieve a 
given F1 score. Since the Random and Active Learning sets were constructed from the same finite pool of data, they were 
guaranteed to have identical performance at the endpoints, so the effect is perhaps even underestimated. In practice, given a 
practically unlimited pool of unlabeled data, the benefits of active learning might be even larger.

The mean stopping point for the Active Learning set in Figure \ref{f:means_all_algorithms} appears to occur right where it 
would be desired, as performance reaches a plateau. Peak or near-peak performance with Active Learning is achieved with 
less than half the dataset. The mean stopping point for the Random set occurs slightly earlier than might be desired. 
This is likely due to the higher variance in stopping points seen in the individual folds in Figure \ref{f:random_noPA} as 
compared to in Figure \ref{f:closest_noPA}. This in turn is presumably due to the more erratic trajectories of the 
Random folds, whereas the Active Learning folds stabilize right around where the stopping points occur.

The model performance metrics corroborate the early stopping points, with the final model trained on 100\% of the data performing roughly 
equivalently to the SVM model created with less than 50\% of the data at the stopping point (see Figure~\ref{f:means_all_algorithms}).

\section{Confidence Experiments} \label{confidenceExperiments}

Given the large volume of tweets, downstream users may only be able to consume a small proportion of tweets deemed relevant. In addition, consumers will have different preferences for the tradeoff between precision and recall, so it is useful to be able to filter out tweets that the classifier is not able to label with high confidence. 

The default decision rule for a binary SVM-based classifier treats all tweets on one side of the separating hyperplane as members of one class (relevant), and all tweets on the other side as members of the other class (irrelevant). An alternative is to only give labels to tweets that are sufficiently far from the hyperplane and discard as ``uncertain'' all tweets that are close to the hyperplane. For example, if a user only had the capacity to process 1,000 relevant tweets and the classifier returned 25,000 relevant tweets, then the user would wish to select the subset of 1,000 tweets most likely to actually be relevant. One way to do this might be to select the 1,000 tweets farthest from the hyperplane. Another user might not have a fixed capacity for processing relevant tweets but might wish to achieve a certain recall level, in which case that user would set a different distance threshold for discarding or retaining tweets. Using distance thresholds in this fashion only makes sense if there is a positive relationship between distance from the hyperplane and classification accuracy.
 
Here, we examine this relationship by setting different distance thresholds and observing how classifier performance changes when tweets whose scores fall below the threshold are discarded. The model predictions (scores) consist of positive and negative real numbers, with the magnitude indicating the Euclidean distance from the hyperplane and the sign indicating which side of the hyperplane the tweet falls on. We converted the scores for the tweets in the test folds to absolute values and then set a series of thresholds from 0 to 2. The thresholds are set in a non-linear fashion due to the fact that the large majority of the scores fall in the range (-1, 1). For each threshold $T$, tweets for which $\vert$score$\vert$ $< T$ are discarded, and only tweets that fall outside the threshold are used to calculate performance metrics. 

\begin{figure}
	\centering
	\includegraphics[keepaspectratio=true, width=\columnwidth]{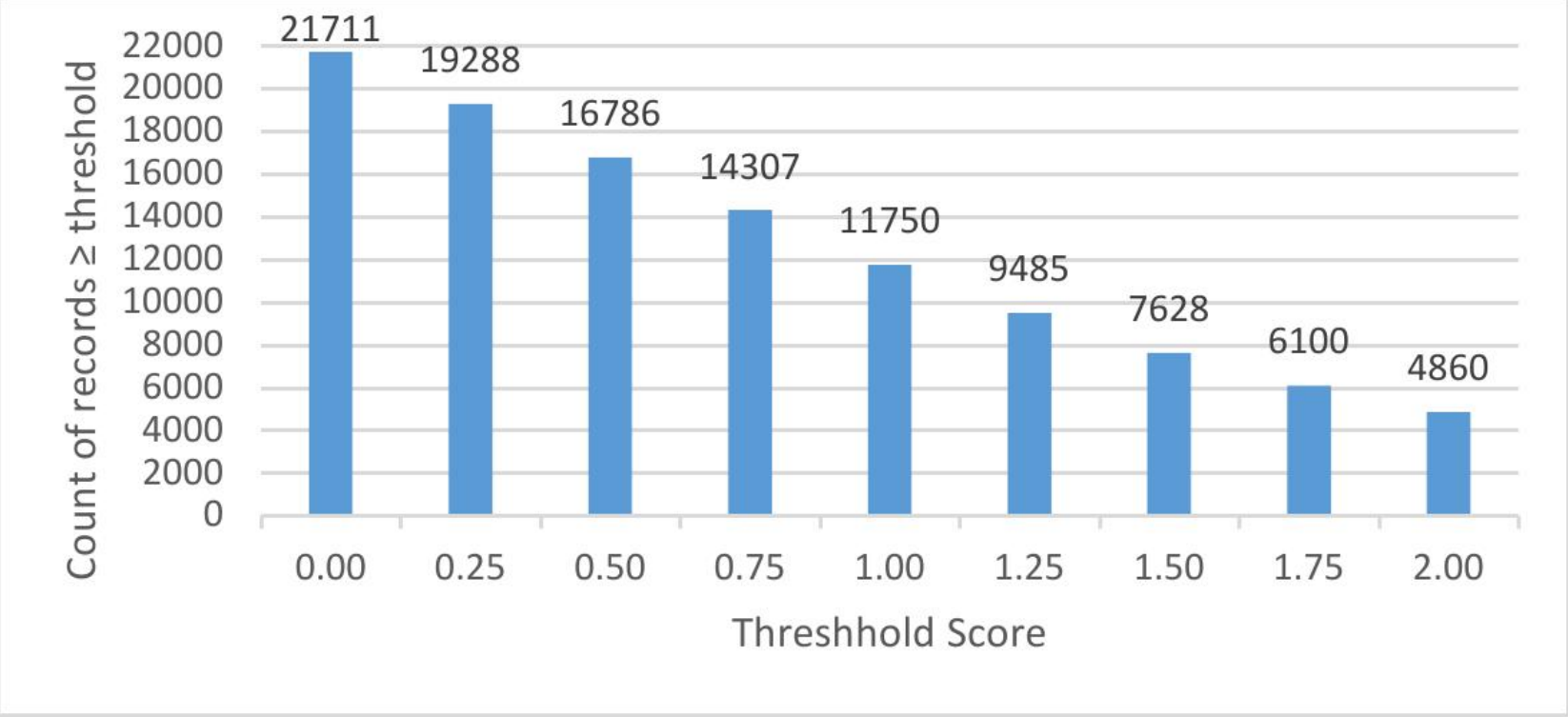}
\caption{ Count of tweets above each threshold }
\label{fig:conf1}
\end{figure}

\begin{figure}
	\centering
	\includegraphics[keepaspectratio=true, width=\columnwidth]{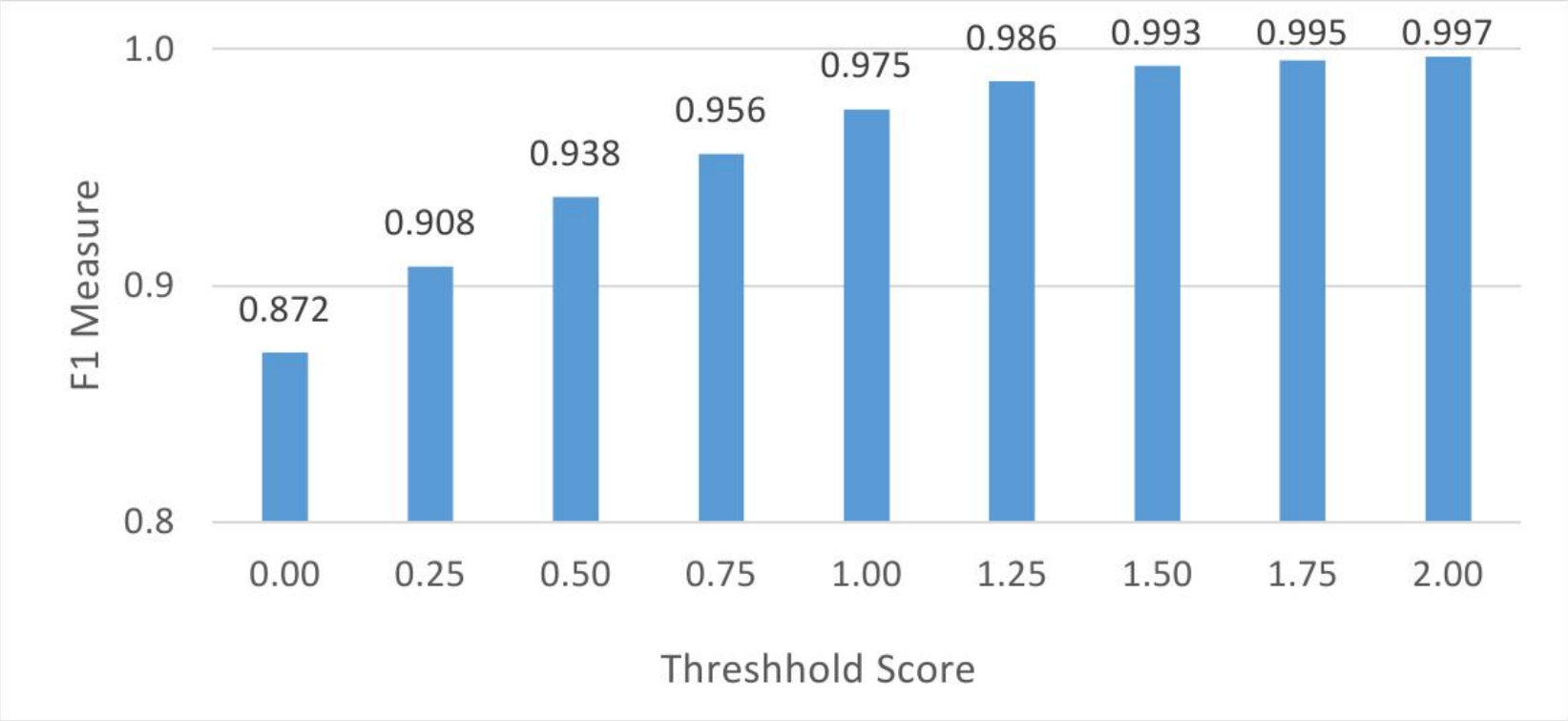}
\caption{ F1 score as a function of threshold }
\label{fig:conf2}
\end{figure}

As the threshold increases, more tweets are discarded, so the count of tweets outside the threshold necessarily decreases (Figure \ref{fig:conf2}). Nevertheless, these tweet counts are high enough to provide robust estimates of F1 at each threshold. Figure \ref{fig:conf1} shows how setting a threshold improves system confidence in the underlying SVM prediction.

Confidence in the classifier's labels of relevant and irrelevant increases rapidly as low magnitude scores are discarded. Even a threshold as low as 0.5 improves the F1 measure by an increment of 0.066, while discarding less than a quarter of the tweets as uncertain. A higher threshold of 1.0 increases the F1 measure to a remarkably high 0.975, albeit almost half of the available data is discarded as uncertain. This tradeoff is desirable in cases where time and resource constraints enable humans to examine only a tiny fraction of the available records. This tradeoff may be relevant to automated systems too, which may hit resource limits as data continues to proliferate.

\begin{figure}
\includegraphics[keepaspectratio=true, width=\columnwidth]{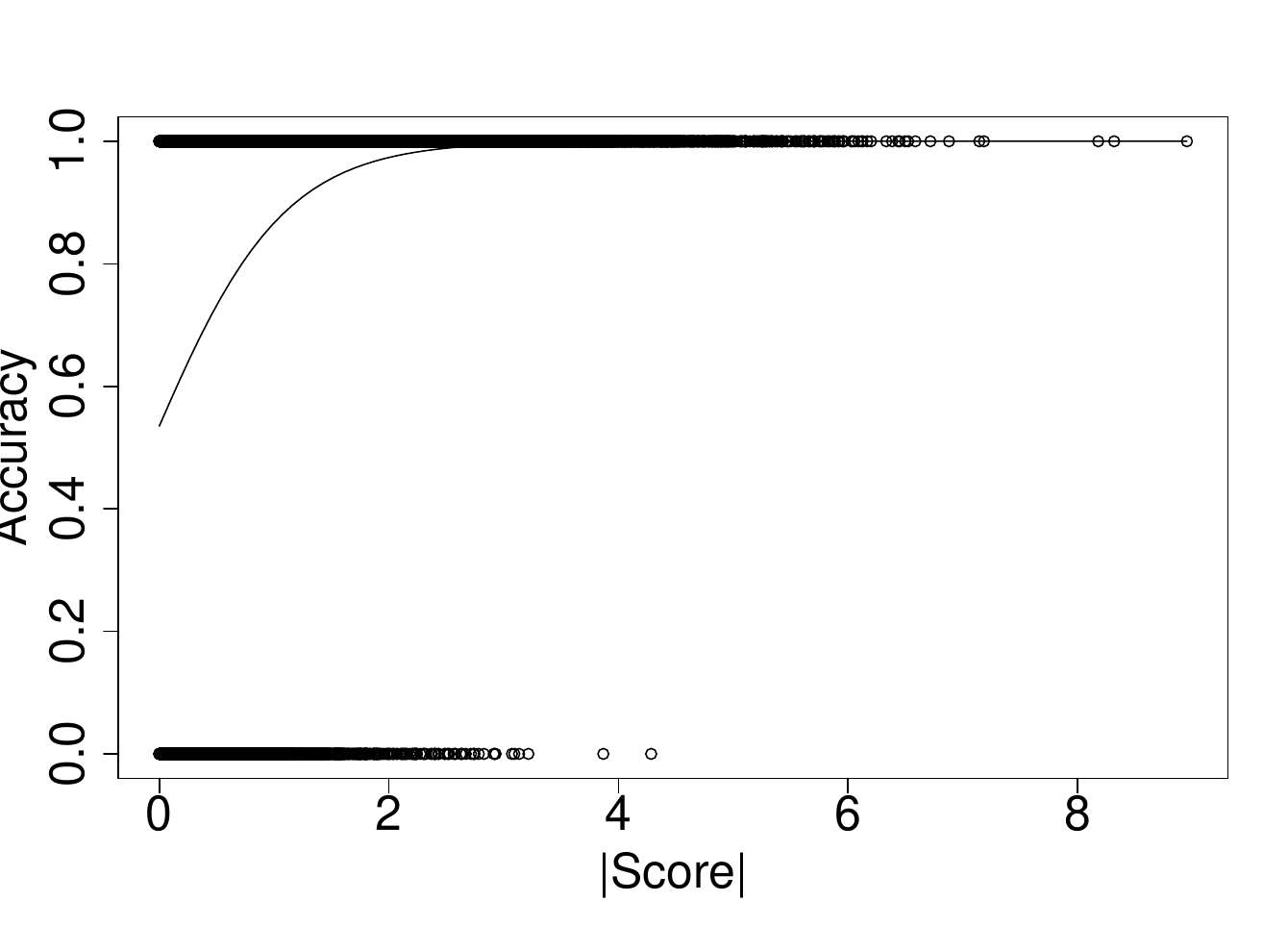}
\caption{Accuracy vs. classifier scores (absolute values) with logistic regression line}
\label{f:confidence_reg}
\vspace{-.5cm}
\end{figure}

\begin{figure}
\includegraphics[keepaspectratio=true, width=\columnwidth]{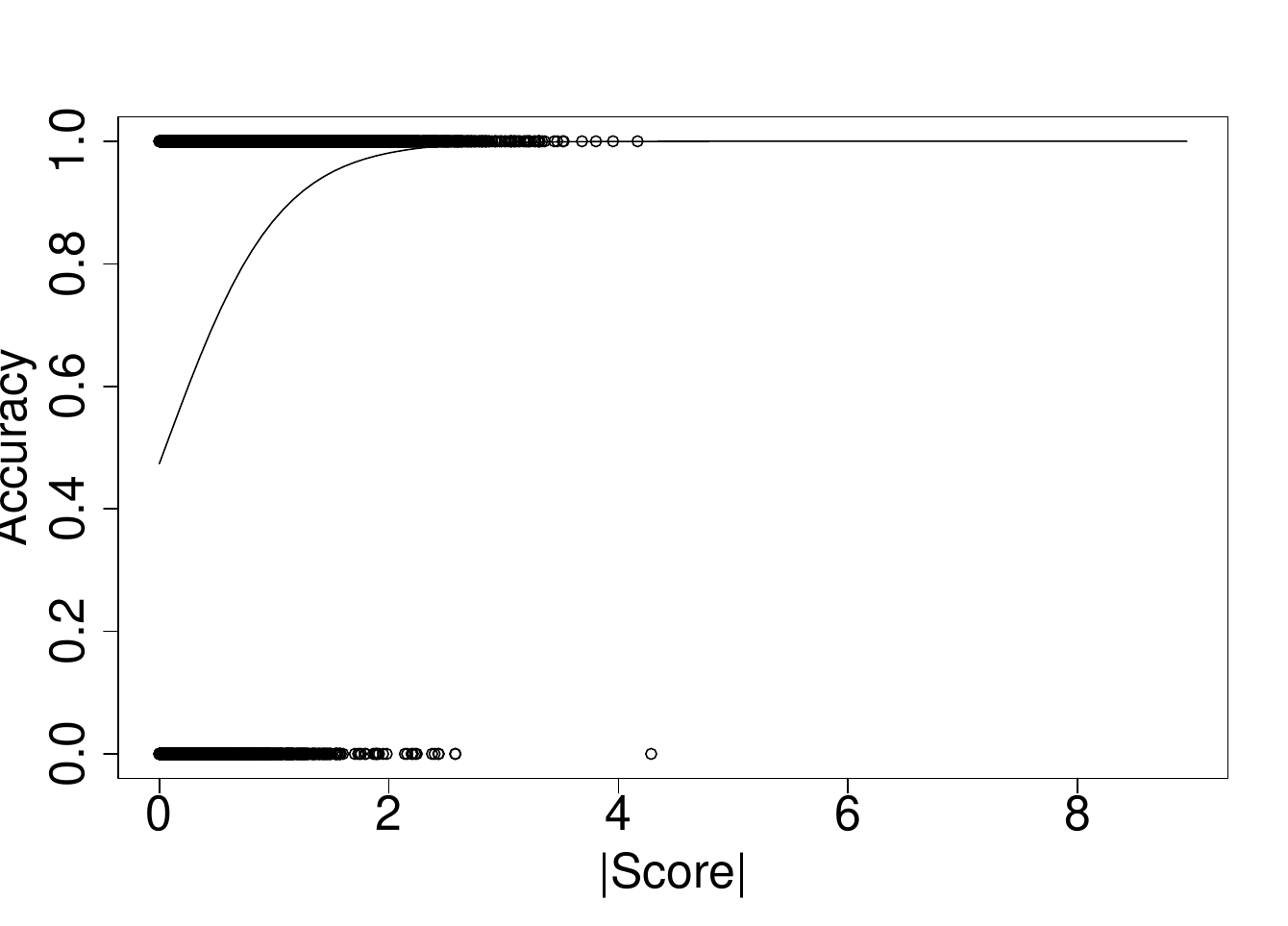}
\caption{Accuracy vs. classifier scores (absolute values) with logistic regression line, negative scores only}
\label{f:confidence_reg_neg}
\vspace{-.5cm}
\end{figure}

\begin{figure}
\includegraphics[keepaspectratio=true, width=\columnwidth]{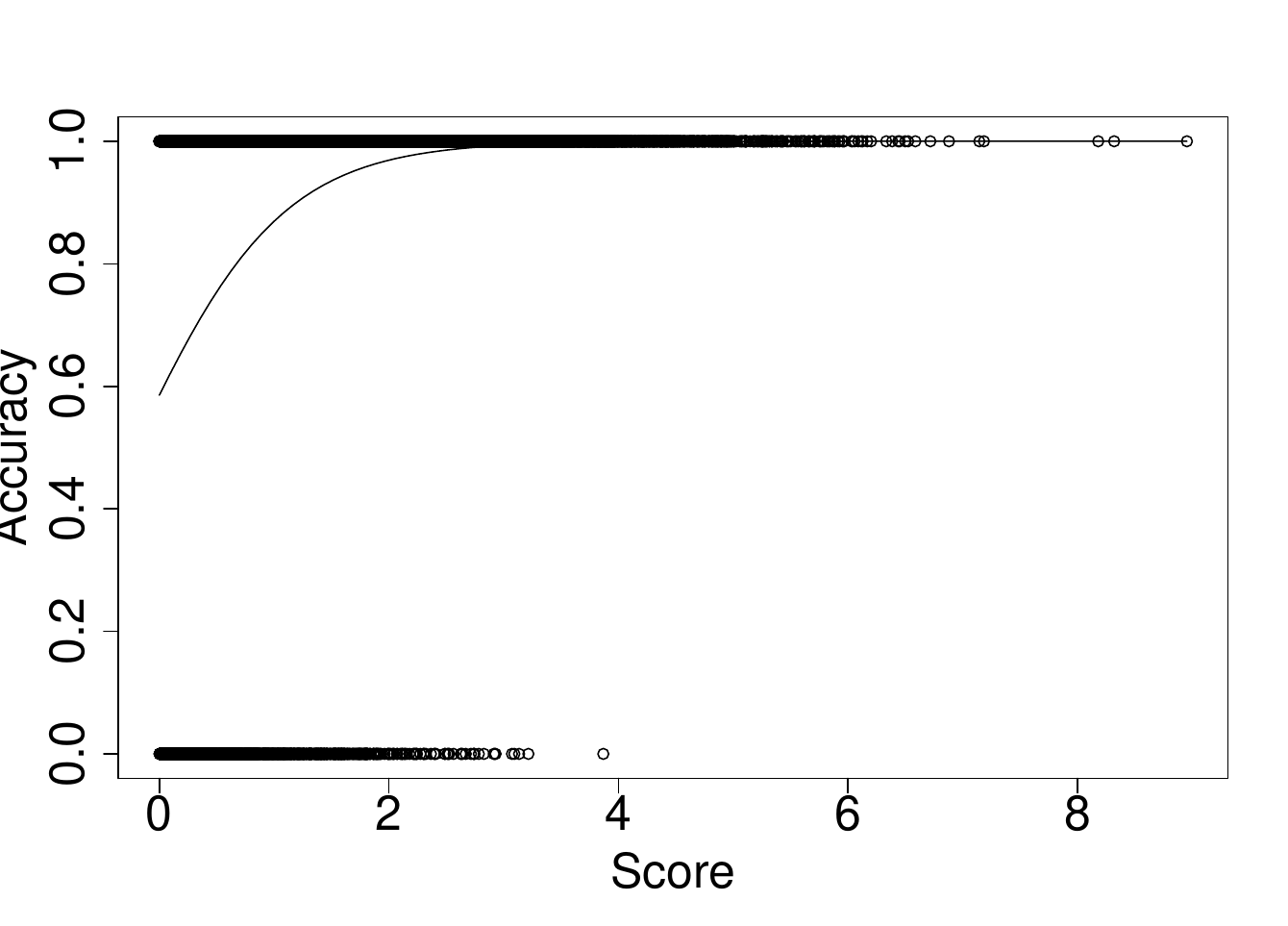}
\caption{Accuracy vs. predictions with logistic regression line, positive scores only}
\label{f:confidence_reg_pos}
\vspace{-.5cm}
\end{figure}

In order to examine the relationship between prediction score and performance in a continuous fashion, we regressed accuracy against the absolute values of the scores. (We used accuracy rather than F1 here, since computing F1 would require scores to be binned.) Figure \ref{f:confidence_reg} shows the results for all scores; Figures \ref{f:confidence_reg_neg} and \ref{f:confidence_reg_pos} show the results for tweets whose raw scores were negative (tweets labeled irrelevant) and positive (tweets labeled relevant), respectively. Wald tests on the regression coefficients were highly significant (\textit{p}'s $< 2^{-16}$), suggesting that accuracy genuinely improves as distance from the hyperplane increases.

The accuracy curve in Figure \ref{f:confidence_reg} mirrors the F1 scores in Figures \ref{fig:conf1}, with the regression curve nearing perfect accuracy around a score of 3. The regression line for the negative scores is sharper than for the positive scores, indicating that negative scores in the range (-2, 0) are slightly more reliable than positive scores in the range (0, 2).
\section{Conclusion} \label{conclusion}

This paper addresses the challenge of anticipating social unrest using Twitter data, focusing on the Arabic-speaking world. In particular, we aimed to create a tool for filtering tweets to provide downstream users with tweets relevant to unrest. Using annotated data, we trained a bag-of-lemmas SVM classifier to identify tweets relevant to unrest with a high degree of reliability. We investigated the relationship between training data size and performance, finding that fewer than 10,000 tweets are sufficient to achieve peak or near-peak performance. We also found that active learning can provide substantial benefits over random sampling for selecting unlabeled tweets to be annotated, achieving superior performance for a given training data size. We also investigated a stopping algorithm designed to identify the point at which further annotation is unnecessary, finding that this algorithm worked well in conjunction with active learning. Finally, we found that higher absolute classifier scores correlate with higher accuracy and F1 scores, and we showed that particular levels of performance can be achieved by setting different thresholds.

The simplest way to identify tweets relevant to a particular topic, such as social unrest, is by means of a whitelist. However, this approach will always be limited by polysemy, as well as by the fact that words related to unrest can appear in many contexts. For example, the word ``strike'' in English can refer to an organized protest, a military attack, a person lighting a match, or a strike in baseball, among other meanings. The word ``military'' can occur in sentences related to combat as well as sentences related to parades, national finances, etc. 

Previous researchers have noted that keyword filters are not sufficient for identifying tweets relevant to social unrest \cite{compton2014}. The whitelist we used to collect our data yielded an estimated relevance rate of 62\%. Our model provides substantial improvement over this baseline, with overall precision, recall, and F1 scores of 0.87, 0.88, and 0.88. The performance can be improved to arbitrarily high levels on subsets of the data by simply adjusting the classifier score threshold and only considering tweets with scores beyond that threshold. A near-perfect F1 score of 0.975 only required discarding half of the data, so a downstream user could obtain a large volume of relevant tweets for further analysis with relatively little cost. Because the whitelist we used contained high-frequency words in Modern Standard Arabic that were designed not to be specific to any particular place, conflict, of political figure, this model is expected to be useful for identifying tweets related to unrest with respect to future conflicts anywhere in the Arabic-speaking world.

\section*{Acknowledgment}

We gratefully acknowledge that a portion of this work was conducted while we were employed at the University of Maryland Center for Advanced Study of Language. This material is based upon work supported, in whole or in part, with funding from the United States Government. Any opinions, findings and conclusions or recommendations expressed in this material are those of the author(s) and do not necessarily reflect the views of the University of Maryland, College Park and/or any agency or entity of the United States Government.

\bibliographystyle{IEEEtran}
\bibliography{paper}

% Generated by IEEEtran.bst, version: 1.14 (2015/08/26)
\begin{thebibliography}{10}
\providecommand{\url}[1]{#1}
\csname url@samestyle\endcsname
\providecommand{\newblock}{\relax}
\providecommand{\bibinfo}[2]{#2}
\providecommand{\BIBentrySTDinterwordspacing}{\spaceskip=0pt\relax}
\providecommand{\BIBentryALTinterwordstretchfactor}{4}
\providecommand{\BIBentryALTinterwordspacing}{\spaceskip=\fontdimen2\font plus
\BIBentryALTinterwordstretchfactor\fontdimen3\font minus
  \fontdimen4\font\relax}
\providecommand{\BIBforeignlanguage}[2]{{%
\expandafter\ifx\csname l@#1\endcsname\relax
\typeout{** WARNING: IEEEtran.bst: No hyphenation pattern has been}%
\typeout{** loaded for the language `#1'. Using the pattern for}%
\typeout{** the default language instead.}%
\else
\language=\csname l@#1\endcsname
\fi
#2}}
\providecommand{\BIBdecl}{\relax}
\BIBdecl

\bibitem{compton2014}
\BIBentryALTinterwordspacing
R.~Compton, C.~Lee, J.~Xu, L.~Artieda-Moncada, T.-C. Lu, L.~D. Silva, and
  M.~Macy, ``Using publicly visible social media to build detailed forecasts of
  civil unrest,'' \emph{Security Informatics}, vol.~3, no.~1, pp. 1--10, 2014.
  [Online]. Available: \url{http://dx.doi.org/10.1186/s13388-014-0004-6}
\BIBentrySTDinterwordspacing

\bibitem{muthiah2015}
S.~Muthiah, B.~Huang, J.~Arredondo, D.~Mares, L.~Getoor, G.~Katz, and
  N.~Ramakrishnan, ``Planned protest modeling in news and social media.'' in
  \emph{AAAI}, 2015, pp. 3920--3927.

\bibitem{ramakrishnan2014}
N.~Ramakrishnan, P.~Butler, S.~Muthiah, N.~Self, R.~Khandpur, P.~Saraf,
  W.~Wang, J.~Cadena, A.~Vullikanti, G.~Korkmaz, C.~Kuhlman, A.~Marathe,
  L.~Zhao, T.~Hua, F.~Chen, C.~T. Lu, B.~Huang, A.~Srinivasan, K.~Trinh,
  L.~Getoor, G.~Katz, A.~Doyle, C.~Ackermann, I.~Zavorin, J.~Ford, K.~Summers,
  Y.~Fayed, J.~Arredondo, D.~Gupta, and D.~Mares, ``'beating the news' with
  embers: Forecasting civil unrest using open source indicators,'' in
  \emph{Proceedings of the 20th ACM SIGKDD International Conference on
  Knowledge Discovery and Data Mining}, ser. KDD '14.\hskip 1em plus 0.5em
  minus 0.4em\relax New York, NY, USA: ACM, 2014, pp. 1799--1808.

\bibitem{campbell2011}
\BIBentryALTinterwordspacing
D.~Campbell, A.~Van~Zee, and D.~Wassink, \emph{Egypt Unshackled: Using Social
  Media to @\#:) the System}.\hskip 1em plus 0.5em minus 0.4em\relax Cambria
  Books, 2011. [Online]. Available:
  \url{https://books.google.com/books?id=juYwYAAACAAJ}
\BIBentrySTDinterwordspacing

\bibitem{xu2014}
\BIBentryALTinterwordspacing
J.~Xu, T.-C. Lu, R.~Compton, and D.~Allen, \emph{Civil Unrest Prediction: A
  Tumblr-Based Exploration}.\hskip 1em plus 0.5em minus 0.4em\relax Cham:
  Springer International Publishing, 2014, pp. 403--411. [Online]. Available:
  \url{http://dx.doi.org/10.1007/978-3-319-05579-4_49}
\BIBentrySTDinterwordspacing

\bibitem{boecking2015}
\BIBentryALTinterwordspacing
B.~Boecking, M.~Hall, and J.~Schneider, ``Event prediction with learning
  algorithms—a study of events surrounding the egyptian revolution of 2011 on
  the basis of micro blog data,'' \emph{Policy \& Internet}, vol.~7, no.~2, pp.
  159--184, 2015. [Online]. Available: \url{http://dx.doi.org/10.1002/poi3.89}
\BIBentrySTDinterwordspacing

\bibitem{hua2013}
T.~Hua, C.-T. Lu, N.~Ramakrishnan, and K.~Summers, ``Analyzing civil unrest
  through social media,'' \emph{Computer}, vol.~46, pp. 80--84, 2013.

\bibitem{morstatter2014}
F.~Morstatter, N.~Lubold, H.~Pon-Barry, J.~Pfeffer, and H.~Liu, ``Finding
  eyewitness tweets during crises,'' in \emph{Proceedings of the ACL 2014
  Workshop on Language Technologies and Computational Social Science}.\hskip
  1em plus 0.5em minus 0.4em\relax Baltimore, MD, USA: Association for
  Computational Linguistics, June 2014, pp. 23--27.

\bibitem{jumadi2016}
\BIBentryALTinterwordspacing
Jumadi, D.~S. Maylawati, B.~Subaeki, and T.~Ridwan, ``Opinion mining on twitter
  microblogging using support vector machine: Public opinion about state
  islamic university of bandung,'' in \emph{2016 4th International Conference
  on Cyber and IT Service Management}.\hskip 1em plus 0.5em minus 0.4em\relax
  IEEE, April 2016, pp. 1--6. [Online]. Available:
  \url{http://ieeexplore.ieee.org/document/7577569/}
\BIBentrySTDinterwordspacing

\bibitem{conover2011}
M.~D. Conover, B.~Gon{\c{c}}alves, J.~Ratkiewicz, A.~Flammini, and F.~Menczer,
  ``{Predicting the political alignment of twitter users},'' in
  \emph{Proceedings - 2011 IEEE International Conference on Privacy, Security,
  Risk and Trust and IEEE International Conference on Social Computing,
  PASSAT/SocialCom 2011}, 2011, pp. 192--199.

\bibitem{dilrukshi2013}
I.~Dilrukshi, K.~{De Zoysa}, and A.~Caldera, ``{Twitter news classification
  using SVM},'' \emph{Proceedings of the 8th International Conference on
  Computer Science and Education, ICCSE 2013}, no. Iccse, pp. 287--291, 2013.

\bibitem{aramaki2011}
\BIBentryALTinterwordspacing
E.~Aramaki, ``{Twitter Catches The Flu : Detecting Influenza Epidemics using
  Twitter The University of Tokyo The University of Tokyo National Institute
  of},'' \emph{Computational Linguistics}, vol. 2011, pp. 1568--1576, 2011.
  [Online]. Available: \url{http://www.aclweb.org/anthology/D11-1145}
\BIBentrySTDinterwordspacing

\bibitem{cohen1960}
J.~Cohen, ``A coefficient of agreement for nominal scales,'' \emph{Educational
  and Psychological Measurement}, vol.~20, pp. 37--46, apr 1960.

\bibitem{fleiss1971}
J.~L. Fleiss, ``Measuring nominal scale agreement among many raters,''
  \emph{Psychological Bulletin}, vol.~76, pp. 378--382, nov 1971.

\bibitem{shrout1979}
P.~E. Shrout and J.~L. Fleiss, ``Intraclass correlations: Uses in assessing
  rater reliability,'' \emph{Psychological Bulletin}, vol.~86, pp. 420--428,
  1979.

\bibitem{whitlock2015}
T.~Whitlock, ``Emoji unicode tables,''
  \url{http://apps.timwhitlock.info/emoji/tables/unicode}, 2015, accessed:
  2016-02-29.

\bibitem{pasha2014}
A.~Pasha, M.~Al-Badrashiny, M.~Diab, A.~E. Kholy, R.~Eskander, N.~Habash,
  M.~Pooleery, O.~Rambow, and R.~Roth, ``\BIBforeignlanguage{english}{Madamira:
  A fast, comprehensive tool for morphological analysis and disambiguation of
  arabic},'' in \emph{\BIBforeignlanguage{english}{Proceedings of the Ninth
  International Conference on Language Resources and Evaluation
  (LREC'14)}}.\hskip 1em plus 0.5em minus 0.4em\relax Reykjavik, Iceland:
  European Language Resources Association (ELRA), May 2014, pp. 1094--1101.

\bibitem{hall2009}
M.~Hall, E.~Frank, G.~Holmes, B.~Pfahringer, P.~Reutemann, and I.~H. Witten,
  ``The weka data mining software: An update,'' \emph{SIGKDD Explorations},
  vol.~11, no.~1, pp. 10--18, Nov. 2009.

\bibitem{joachims1999}
T.~Joachims, ``Making large-scale {SVM} learning practical,'' in \emph{Advances
  in Kernel Methods -- Support Vector Learning}, 1999, pp. 169--184.

\bibitem{bloodgood2009a}
M.~Bloodgood and K.~Vijay-Shanker, ``Taking into account the differences
  between actively and passively acquired data: The case of active learning
  with support vector machines for imbalanced datasets,'' in \emph{Proceedings
  of Human Language Technologies: The 2009 Annual Conference of the North
  American Chapter of the Association for Computational Linguistics, Companion
  Volume: Short Papers}.\hskip 1em plus 0.5em minus 0.4em\relax Boulder,
  Colorado: Association for Computational Linguistics, June 2009, pp. 137--140.

\bibitem{campbell2000}
C.~Campbell, N.~Cristianini, and A.~J. Smola, ``Query learning with large
  margin classifiers,'' in \emph{Proceedings of the 17th International
  Conference on Machine Learning (ICML)}, 2000, pp. 111--118.

\bibitem{schohn2000}
G.~Schohn and D.~Cohn, ``Less is more: {A}ctive learning with support vector
  machines,'' in \emph{Proceedings of the 17th International Conference on
  Machine Learning (ICML)}, 2000, pp. 839--846.

\bibitem{tong2002}
S.~Tong and D.~Koller, ``Support vector machine active learning with
  applications to text classification,'' \emph{Journal of Machine Learning
  Research (JMLR)}, vol.~2, pp. 45--66, 2002.

\bibitem{bloodgood2009b}
M.~Bloodgood and K.~Vijay-Shanker, ``A method for stopping active learning
  based on stabilizing predictions and the need for user-adjustable stopping,''
  in \emph{Proceedings of the Thirteenth Conference on Computational Natural
  Language Learning (CoNLL-2009)}.\hskip 1em plus 0.5em minus 0.4em\relax
  Boulder, Colorado: Association for Computational Linguistics, June 2009, pp.
  39--47.

\bibitem{bloodgood2013}
M.~Bloodgood and J.~Grothendieck, ``Analysis of stopping active learning based
  on stabilizing predictions,'' in \emph{Proceedings of the Seventeenth
  Conference on Computational Natural Language Learning}.\hskip 1em plus 0.5em
  minus 0.4em\relax Sofia, Bulgaria: Association for Computational Linguistics,
  August 2013, pp. 10--19.

\end{thebibliography}

\end{document}